\documentclass{mva_style}
\usepackage[dvipdfmx]{graphicx}
\usepackage{support-caption}
\usepackage{subcaption}
\usepackage{adjustbox}
\usepackage{booktabs}
\usepackage{multirow}
\usepackage{xurl}
\usepackage{amsmath}
\usepackage{pifont}
\usepackage{array} 
\usepackage{float} 

\newcolumntype{C}[1]{>{\centering\arraybackslash}p{#1}}

\usepackage{url}

\usepackage[breaklinks=true,bookmarks=true,colorlinks=true]{hyperref}
\renewcommand{\footnotesize}{\fontsize{7pt}{9pt}\selectfont}

\finalcopy 

\begin{document}
\title{MVA 2025 Small Multi-Object Tracking for Spotting Birds Challenge: Dataset, Methods, and Results}

\author{
Yuki Kondo$^{1, \dag}$ \and Norimichi Ukita$^{2, \dag}$ \and Riku Kanayama$^{2, \dag}$ \and Yuki Yoshida$^{2, \dag}$ \and Takayuki Yamaguchi$^{3, \dag}$ \and
Xiang Yu$^4$ \and Guang Liang$^4$ \and Xinyao Liu$^5$ \and
Guan-Zhang Wang$^6$ \and Wei-Ta Chu$^6$ \and
Bing-Cheng Chuang$^7$ \and Jia-Hua Lee$^7$ \and Pin-Tseng Kuo$^7$ \and I-Hsuan Chu$^7$ \and Yi-Shein Hsiao$^7$ \and Cheng-Han Wu$^7$ \and Po-Yi Wu$^8$ \and Jui-Chien Tsou$^8$ \and Hsuan-Chi Liu$^8$ \and Chun-Yi Lee$^8$ \and Yuan-Fu Yang$^9$ \and
Kosuke Shigematsu$^{10}$ \and Asuka Shin$^{10}$ \and
Ba Tran$^{11}$ \and\\
\small $^1$Toyota Motor Corporation, 
$^2$Toyota Technological Institute, 
$^3$Iwate Prefecture Coastal Regional Development Bureau,\\
\small $^4$Nanjing University,
\small $^5$University of Science and Technology of China,
$^6$National Cheng Kung University,\\
\small $^7$National Tsing Hua University,
$^8$National Taiwan University, $^9$National Yang Ming Chiao Tung University,\\
\small $^{10}$National Institute of Technology, Oita College,
$^{11}$Axelspace Corporation
}

\maketitle

\section*{\centering Abstract}{
\noindent Small Multi-Object Tracking (SMOT) is particularly challenging when targets occupy only a few dozen pixels, rendering detection and appearance-based association unreliable. Building on the success of the MVA2023 SOD4SB challenge, this paper introduces the SMOT4SB challenge, which leverages temporal information to address limitations of single-frame detection. Our three main contributions are: (1) the SMOT4SB dataset, consisting of 211 UAV video sequences with 108,192 annotated frames under diverse real-world conditions, designed to capture motion entanglement where both camera and targets move freely in 3D; (2) SO-HOTA, a novel metric combining Dot Distance with HOTA to mitigate the sensitivity of IoU-based metrics to small displacements; and (3) a competitive MVA2025 challenge with 78 participants and 308 submissions, where the winning method achieved a 5.1× improvement over the baseline. This work lays a foundation for advancing SMOT in UAV scenarios with applications in bird strike avoidance, agriculture, fisheries, and ecological monitoring.
}

{\let\thefootnote\relax\footnotetext{%
\hspace{-5mm}  \textsuperscript{\dag}~indicates the organizers of the Challenge. The other authors participated in the challenge. Appendix~\ref{sec:apd:team} contains the authors' team names and affiliations.
}
}

\begin{figure}[t]
  \centering
  \includegraphics[width=\linewidth]{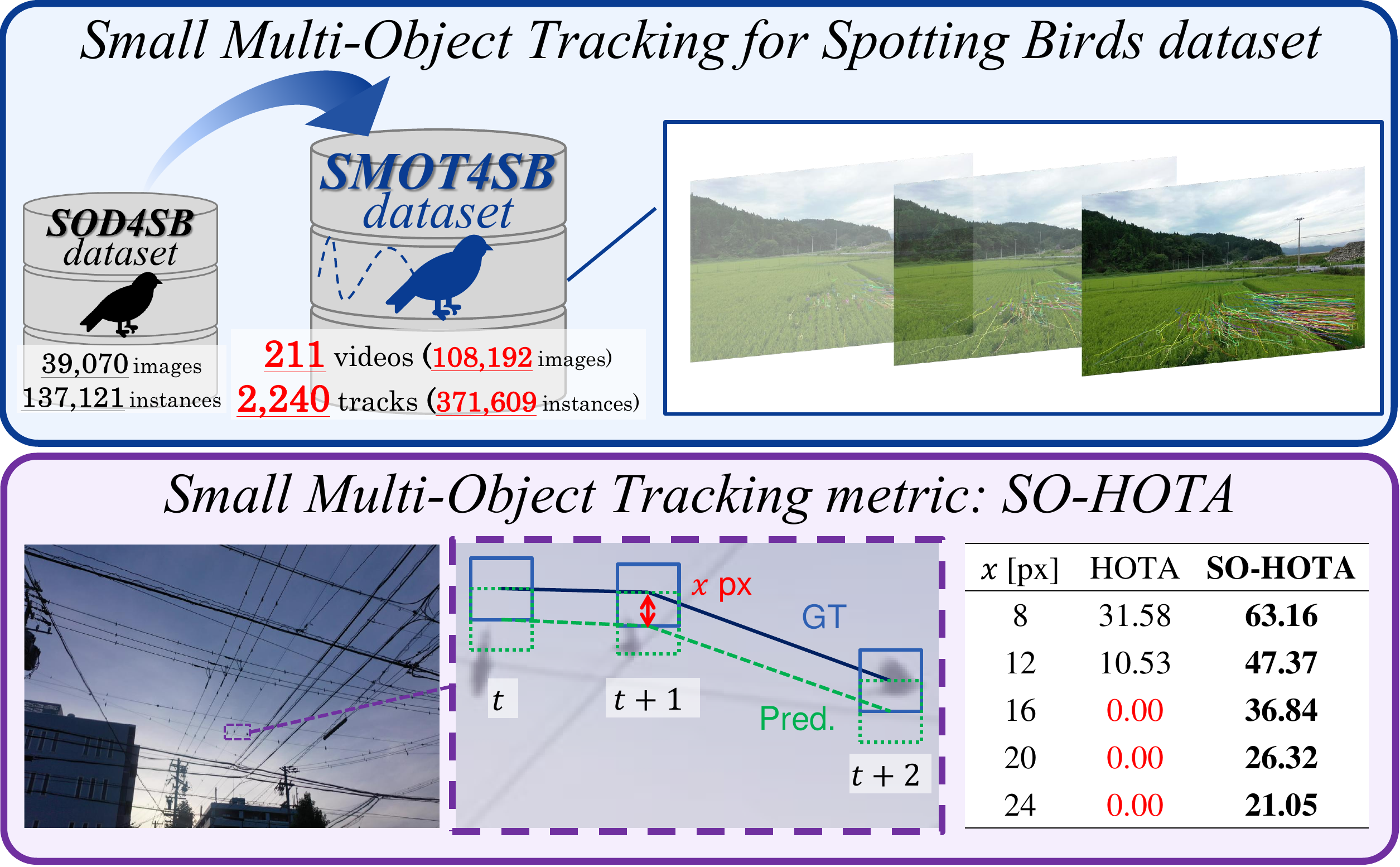}
  \caption{Overview of the SMOT4SB dataset and evaluation metric.
(Top) SMOT4SB extends the SOD4SB dataset~\cite{mva2023_sod_challenge} from SOD to SMOT in UAV scenarios, with 211 videos and 2,240 annotated tracks. (Bottom) We introduce SO-HOTA, which incorporates DotD~\cite{xu2021dot} into the HOTA framework~\cite{luiten2021hota} to reduce over-sensitivity to minor localization errors. The comparison assumes a vertical displacement of $x$ px from the ground truth, with a fixed bounding box size of 16 px. While conventional HOTA scores drop sharply to zero under minor displacements, SO-HOTA provides a more stable and discriminative evaluation for small object tracking.}
  \label{fig:teaser}
\end{figure}

\section{Introduction}

Multi-Object Tracking (MOT) represents a fundamental computer vision task that extends object detection to temporal sequences, enabling continuous identification and localization of multiple target objects across video frames~\cite{ciaparrone2020deep,dendorfer2020motchallenge}. Significant progress has been achieved in general MOT scenarios through advances in detection-based tracking~\cite{sort_2016,deepsort_2017}, joint detection and tracking~\cite{jde_2020,fairmot_2021}, and transformer-based approaches~\cite{motr_2022,trackformer_2022}.

Although these developments have improved overall MOT performance, Small Multi-Object Tracking (SMOT) remains a particularly challenging frontier. SMOT introduces unique technical complexities that distinguish it from conventional MOT frameworks. Objects spanning merely a few pixels suffer from severe information scarcity, making appearance-based association unreliable~\cite{liu2019aggregation_signature,yao2023folt}. Background clutter and rapid scale changes exacerbate detection consistency problems, while traditional tracking methods exhibit significant performance degradation when object sizes fall below $32 \times 32$ pixels as established by MS COCO standards~\cite{COCO_ECCV2014}.

As a preliminary step toward SMOT, the organizers previously addressed the problem of small object detection by organizing the MVA2023 Small Object Detection Challenge for Spotting Birds (SOD4SB)~\cite{mva2023_sod_challenge}. Focused on UAV-based wild bird detection, SOD4SB demonstrated the feasibility of detecting small, distant, and fast-moving objects under real-world environmental conditions. With 223 participants, the challenge not only advanced small object detection but also suggested promising applications in autonomous UAV systems for avoiding bird strikes~\cite{fujii2021distant}, protection against bird-related damage in fields, aviation, and fisheries~\cite{dehaven1981estimating,hedayati2015bird,spanier1980use}, and ecological monitoring through automated bird population surveys~\cite{DBLP:conf/igarss/OgawaLTHKM21}.

However, SOD4SB's constraint to single-frame detection fundamentally limited both the detection performance and practical system robustness. When appearance features are insufficient due to small object regions in each frame, detection becomes unreliable under challenging conditions such as occlusions, motion blur, and background clutter. This limitation directly constrains the reliability of subsequent tracking processes, as inconsistent detection forms a bottleneck for temporal association and identity preservation in SMOT scenarios.

Building upon this foundation, the organizers extend SOD4SB to Small Multi-Object Tracking for Spotting Birds (SMOT4SB) to address the fundamental limitations of single-frame detection. SMOT4SB explores how temporal information can compensate for the information scarcity of small objects by integrating spatial and motion cues. This hypothesis is supported by recent work showing that motion-guided features from aligned video frames significantly improve small object detection when appearance cues are weak or missing~\cite{yang2023video}, consistent with the biological role of the dorsal stream in motion perception~\cite{choi2020proposal}.

Moreover, SMOT4SB introduces a substantially more challenging and diverse computer vision problem setting that promotes methodological innovation. Unlike most conventional MOT and SMOT scenarios~\cite{dendorfer2020motchallenge,chavdarova2018wildtrack,rezaeinia2016social,visdrone} that assume fixed cameras or objects constrained to planar motion, SMOT4SB presents a scenario where both the camera platform and the target objects move freely in three-dimensional space. This scenario requires consideration of what the organizers term \emph{motion entanglement}—the intertwined state of ego-motion and subject motion that must be disentangled for effective tracking. Furthermore, the flocking behavior exhibited by some bird species introduces fascinating opportunities to model interactions between individual and collective motion dynamics, presenting compelling research questions from a Social Force Model perspective~\cite{helbing1995social,yamaguchi2011you}.

To realize this challenging problem setting, the organizers establish the research infrastructure necessary for the systematic exploration of SMOT4SB. While standard MOT benefits from well-established datasets and evaluation protocols, the unique characteristics of tracking small birds from UAV platforms require dedicated research foundations. The combination of extreme object scales, \emph{motion entanglement}, and dynamic 3D environments presents distinct challenges that existing MOT resources cannot adequately address. Additionally, while SMOT research has largely adopted standard MOT evaluation metrics, the specific characteristics of small objects warrant reconsideration of these evaluation approaches. 

This work contributes to building a comprehensive research infrastructure and advancing algorithmic development through the following three key components:
\begin{enumerate}
    \item \textbf{SMOT4SB Dataset}: The organizers introduce a large-scale dataset comprising 211 video sequences with 183,192 annotated frames, representing the first dataset specifically designed for small object tracking in UAV scenarios (Fig.~\ref{fig:teaser}).
  
    \item \textbf{SO-HOTA Metric}: The organizers propose Small Object Higher Order Tracking Accuracy (SO-HOTA), the first evaluation metric specifically tailored for small object tracking tasks, which integrates the Dot Distance (DotD) measure~\cite{xu2021dot} with the Higher Order Tracking Accuracy HOTA metric~\cite{luiten2021hota} to address the excessive sensitivity of IoU-based metrics to small object displacement (Fig.~\ref{fig:teaser}).
  
    \item \textbf{Comprehensive Benchmark}: The organizers provide a competitive challenge with 78 participants and 308 submissions to explore technical approaches for the SMOT4SB problem setting. The winners present the top-5 methods and their innovations, with the winning approach achieving a $5.1\times$ performance improvement over the baseline (i.e., SO-HOTA score of 50.59). Even after the challenge period, CodaBench~\cite{codabench} remains publicly available for evaluation on the public test subset, as described in Appendix.~\ref{sec:apd:resource}.
\end{enumerate}



\section{Related Work}
\label{sec:related_work}

\subsection{Small Object Detection (SOD)}
SOD has evolved by adapting generic object detectors to address limited appearance cues, using multi-scale feature extraction~\cite{fpn}, attention mechanisms~\cite{attention_sod}, and tailored strategies such as super-resolution~\cite{perceptual_gan,tdsr,rdsp} and data augmentation~\cite{copy_paste_augmentation}. Recent methods have demonstrated improved accuracy through test-time ensemble fusion~\cite{team1_mva2023}, Swin Transformer-based architectures~\cite{happy_day_mva2023}, and frequency-based scale analysis~\cite{BandRe_MVAW2023}.

Benchmark datasets have expanded from domain-specific corpora, including person detection~\cite{yu2020scale}, traffic light recognition~\cite{behrendt2017deep}, and aerial imagery~\cite{wang2021tiny}, to bird-focused detection benchmarks~\cite{yoshihashi2017bird, sun2022airbirds, fujii2021distant, mva2023_sod_challenge}.

Recent evaluation metrics for SOD have evolved along two complementary directions: analyzing performance across object scales and mitigating scale sensitivity in matching. For the former, ASAP~\cite{asap} and its improved variant BandASAP~\cite{BandRe_MVAW2023} provide fine-grained AP evaluation using scale-specific weighting. For the latter, DotD~\cite{xu2021dot} evaluates normalized center-point distance instead of spatial overlap, offering a more stable and scale-insensitive alternative, particularly effective for SOD.

Despite these advances, current SOD methods remain fundamentally constrained by single-frame analysis, which is unable to leverage temporal consistency and motion patterns that could enhance the recognition of objects with minimal visual information.

\begin{figure*}[t]
  \begin{center}
    \includegraphics[width=\linewidth]{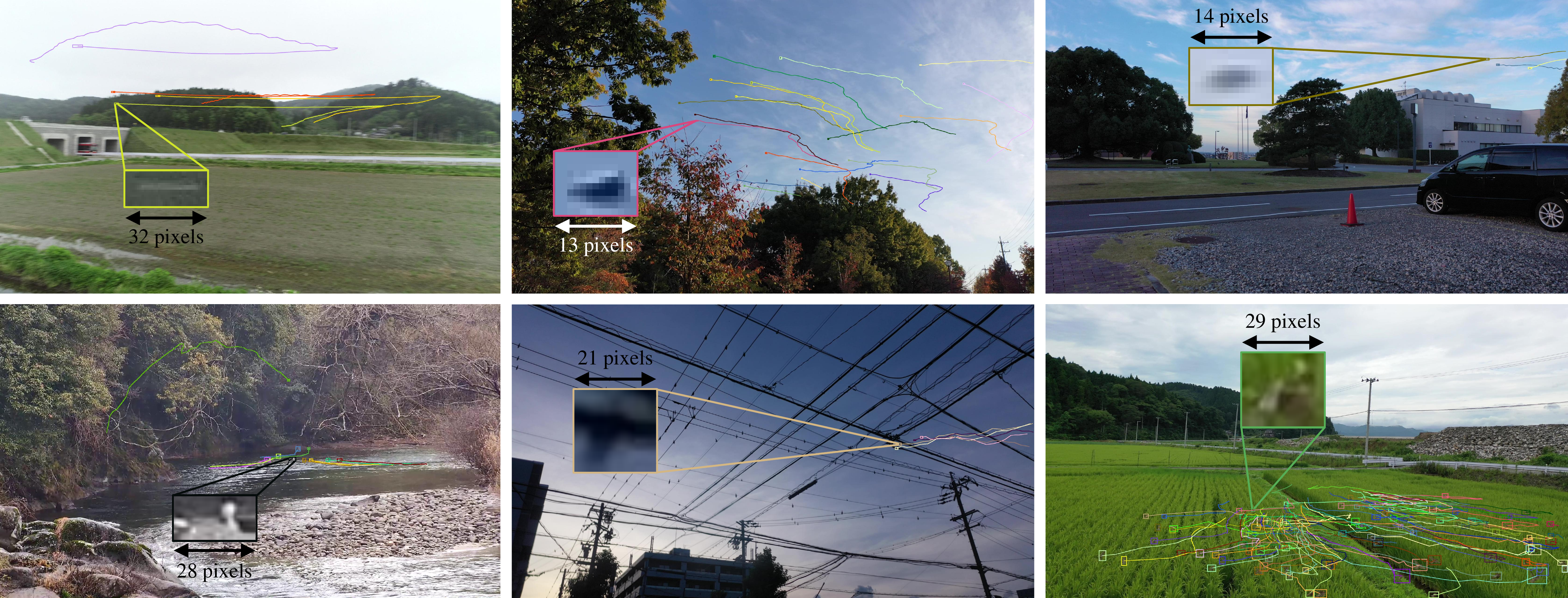}
  \end{center}
  \vspace{-3mm}
  \caption{Examples from the SMOT4SB dataset.}
  \label{fig:smot4sb_sample}
\end{figure*}

\subsection{Multi-Object Tracking (MOT)}
MOT has evolved through distinct paradigms: detection-based tracking~\cite{sort_2016,deepsort_2017}, joint detection-tracking optimization~\cite{fairmot_2021}, and end-to-end Transformer frameworks~\cite{motr_2022}. Recent advances include efficient association methods~\cite{bytetrack} and improvements in speed~\cite {modi2024real}.

General datasets for MOT include MOT Challenge series~\cite{leal2015motchallenge,milan2016mot16,dendorfer2020mot20} for pedestrian tracking, focusing on human-scale objects where appearance features provide reliable association cues. 

In terms of evaluation, while MOTA~\cite{bernardin2008evaluating} depends heavily on detection bias through IDF1 Score~\cite{idf1}, HOTA~\cite{luiten2021hota} provides a balanced assessment of detection and association performance and has become the current gold standard for tracking evaluation.

\subsection{Small Multi-Object Tracking (SMOT)}
SMOT poses challenges where appearance features alone are insufficient for association. Recent SMOT methods address this issue with re-identification modules and specialized tracking strategies. Top entries in the VisDrone-MOT challenge~\cite{visdrone}, such as COFE and SOMOT, enhance the tracking performance through coarse-category training and improved appearance modeling. Other approaches include IoU-guided attention for improved localization~\cite{marvasti2020comet}, transformer-based global context modeling~\cite{liu2021aerial}, and multi-level knowledge distillation to enhance small object features~\cite{latot}.

On the dataset side, existing benchmarks such as Small90~\cite{liu2019aggregation_signature}, UAV123~\cite{uav123}, VisDrone~\cite{visdrone}, Campus~\cite{campus}, and UAVDT~\cite{uavdt} mainly target urban environments with constrained motion. In contrast, wildlife tracking scenarios often involve \emph{motion entanglement}, where both the camera and targets move freely in 3D space, yet such conditions remain relatively underexplored in existing research and datasets.

Regarding evaluation, while HOTA~\cite{luiten2021hota} has become the standard for MOT, its IoU-based formulation is overly sensitive to small displacements. This sensitivity limits its suitability for SMOT tasks involving small, fast-moving objects under complex motion.

\begin{table*}[t]
  \centering
  \caption{Comparison of existing datasets for small object tracking from aerial or distant viewpoints. SOT: single object tracking, Frames: total number of frames, IDs: unique tracking identities, $^\dag$: includes UAV.}
  \label{tab:dataset_summary}
  \resizebox{\linewidth}{!}{
  \begin{tabular}{lp{3.4cm}cC{2.4cm}crrrp{2cm}cc}
    \toprule
    Dataset & Scene & Task & Camera Type & Cam. Move & Videos & Frames & IDs & Resolution & Classes & Year \\
    \midrule
    \multirow{2}{*}{\textbf{SMOT4SB}} & Forest / River / Field / Urban / Park & \multirow{2}{*}{MOT} & \multirow{2}{*}{UAV} & \multirow{2}{*}{\ding{51}} & \multirow{2}{*}{211} & \multirow{2}{*}{108,192} & \multirow{2}{*}{2,240} & 1920×1080 / 3840×2160 & \multirow{2}{*}{1} & \multirow{2}{*}{2025} \\
    \midrule
    LaTOT~\cite{latot}    & Diverse (web-sourced) & SOT & Various\textsuperscript{\dag} & \ding{51} & 434 & 217,700 & -- & -- & 48 & 2023 \\
    VISO~\cite{viso2021}    & Satellite & SOT / MOT & Satellite & -- & 47 & 189,943 & 3,711 & 1000×1000 & 4 & 2023 \\
    VisDrone~\cite{visdrone} & Urban & SOT / MOT & UAV & \ding{51} & 288 & 261,908 & 8,491 & 3840×2160 & 10 & 2018 \\
    UAVDT~\cite{uavdt}   & Urban & MOT & UAV & \ding{51} & 50 & 77,819 & 2,700 & 1080×540 & 3 & 2018 \\
    Campus~\cite{campus}  & Campus & MOT & UAV & -- & 60 & 929,499 & 19,564 & 1904×1400 & 6 & 2016 \\
    UAV123~\cite{uav123}  & Urban & SOT & UAV & \ding{51} & 123 & 112,578 & 123 & 1280×720 & 13 & 2016 \\
    \bottomrule
  \end{tabular}
  }
\end{table*}


\section{SMOT4SB Dataset}
\label{sec:dataset}

Building upon the foundation of SOD4SB~\cite{mva2023_sod_challenge}, we introduce the SMOT4SB dataset to address not only the fundamental limitations of single-frame detection in small object scenarios but also the difficulty in multi-object tracking through novel challenges that existing MOT datasets cannot provide.
SMOT4SB offers unique opportunities to explore \emph{motion entanglement} and to model the intricate interactions between individual and collective motion dynamics exhibited by flocking birds.

As shown in Fig.~\ref{fig:smot4sb_sample}, the SMOT4SB dataset presents a comprehensive collection of video sequences featuring small birds captured from UAV platforms across diverse real-world environments.

Table~\ref{tab:dataset_summary} provides a detailed comparison of SMOT4SB with existing small object tracking datasets.
Our dataset distinguishes itself through its comprehensive scale, diverse environmental conditions, and focus on challenging \emph{motion entanglement} scenarios where both camera platform and subjects move freely in three-dimensional space.
The dataset comprises 211 video sequences with 108,192 annotated frames, representing a dataset specifically designed for small object tracking in UAV scenarios.

\subsection{Collection} 

On-drone cameras were used for video collection. The drones used for filming were the DJI Mavic 2 Pro, the DJI Phantom 4 Pro V2.0, and the ProDrone PD4B-M.
The camera captures videos at 30 fps with resolutions of $1,920 \times 1,080$ and $3,840 \times 2,160$ pixels. Temporal frames in the same video sequence are regarded as consecutive frames in this year's challenge.
The videos were captured across various locations, including urban areas, parks, forests, rivers, and agricultural fields, under different weather and lighting conditions.
Birds observed in the sequences include hawks, crows, waterfowl, sparrows, and various small passerine species.

A key characteristic of our dataset is the presence of flocking behavior, where multiple birds exhibit complex collective motion dynamics that create mutual occlusions and challenging association scenarios.
Due to the rapid motion of both birds and UAV platforms, motion blur and background clutter frequently occur, presenting additional challenges for tracking algorithms.
Since most birds were positioned far from the drone during capture, the majority of bird instances qualify as small objects according to the established definition~\cite{COCO_ECCV2014}.

\subsection{Annotations}

The organizers manually annotated each video sequence where any birds are observed.
Trained annotators used VATIC~\cite{vondrick2013efficiently} and CVAT~\cite{cvat} video annotation tools to enclose each bird instance with a bounding box and assign a unique tracking ID.
All annotations were double-checked to ensure quality.
While several types of wild birds are observed in our dataset, all of them are annotated simply as ``bird", as correctly classifying all small bird instances remains challenging even for human annotators.
A key challenge was maintaining consistency in tracking IDs through complex scenarios, such as occlusions, motion blur, and scale variations.
To address this challenge, annotators employed magnification techniques and performed careful frame-by-frame comparison to ensure accurate correspondence between adjacent frames.

The annotations are provided in the COCO format.
In total, the dataset consists of 211 videos (108,192 frames), containing 371,690 bird instances across 2,240 unique tracking IDs.

\begin{table}[t]
  \centering
  \caption{Statistics of each dataset subset. Instances: number of annotated objects.}
  \label{tab:split_summary}
  \resizebox{\linewidth}{!}{
  \begin{tabular}{lcccc}
    \toprule
    Subset & Videos & Frames & Instances & IDs \\
    \midrule
    Train        & 128 & 66,602  & 226,292 & 1,256 \\
    Public test  & 38  & 16,489  & 51,325  & 509   \\
    Private test & 45  & 25,101  & 94,073  & 475   \\
    \midrule
    Total        & 211 & 108,192 & 371,690 & 2,240 \\
    \bottomrule
  \end{tabular}
  }
\end{table}

\subsection{Splitting}

To ensure fair evaluation, the dataset is split at the video level into training, public test, and private test subsets, such that no video appears in multiple splits. Table~\ref{tab:split_summary} summarizes the statistics and availability of each subset. While the training videos and annotations are publicly available, the public test set includes only videos (annotations withheld), and the private test set is fully non-public.

\subsection{Challenge Attributes}\label{subsub:challenge-attribute}

\begin{figure}[H]
    \centering
    \small
    \begin{subfigure}[t]{\linewidth}
        \centering
        \includegraphics[width=\linewidth]{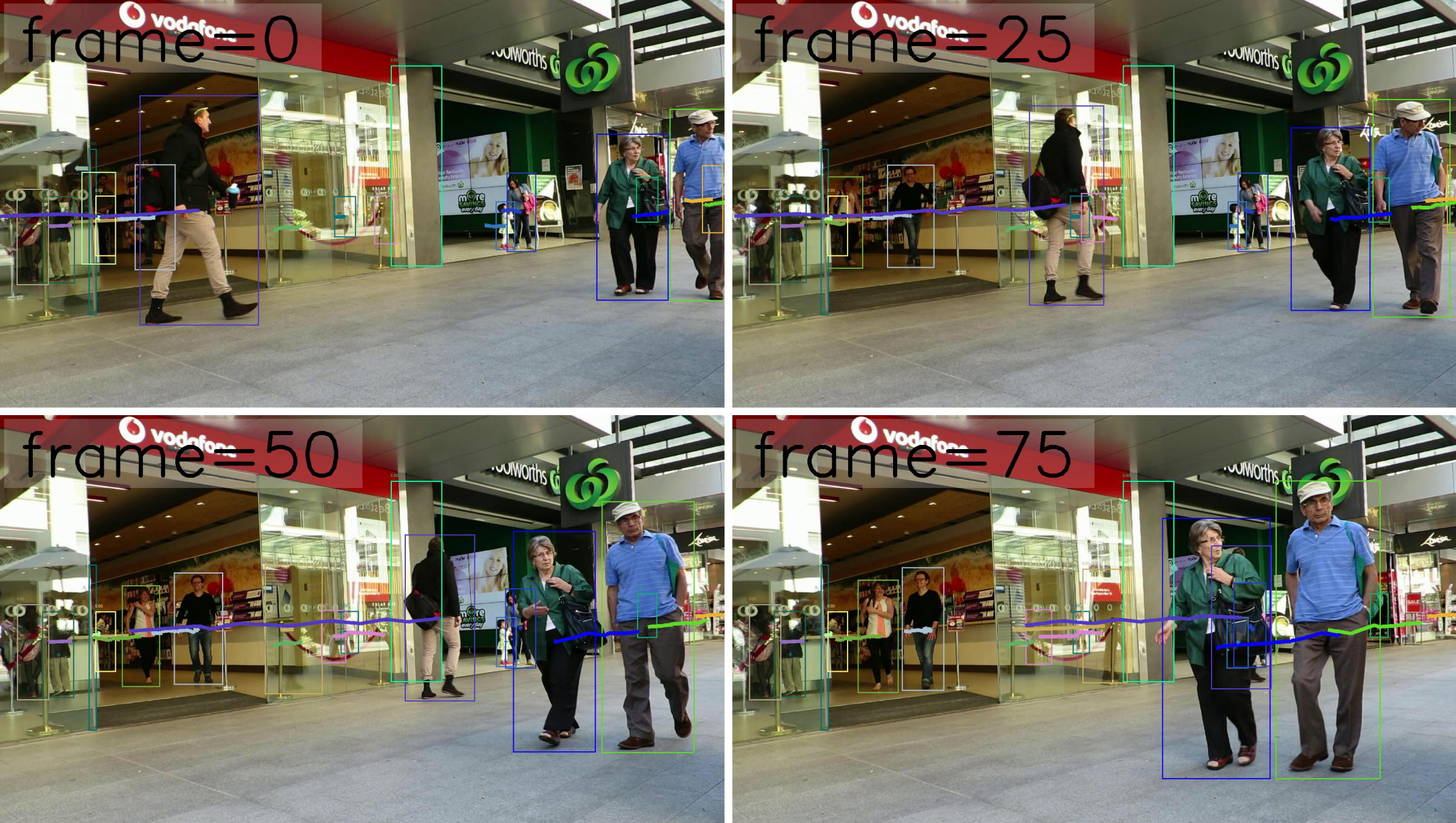}
        \caption{MOT17 dataset~\cite{dendorfer2020motchallenge}}
        \label{fig:mot17_me}
    \end{subfigure}

    \vspace{0.5em}

    \begin{subfigure}[t]{\linewidth}
        \centering
        \includegraphics[width=\linewidth]{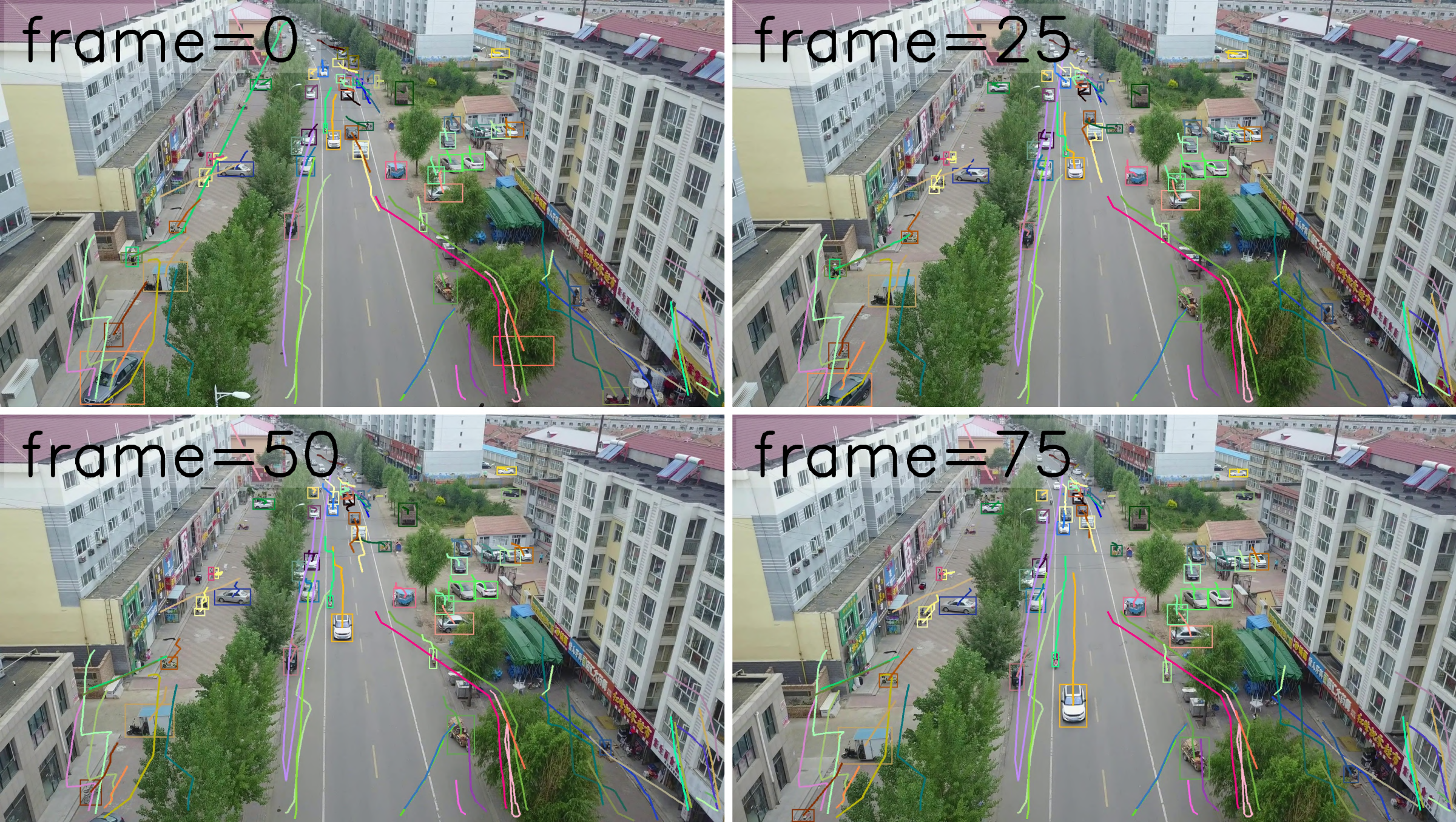}
        \caption{UAVDT dataset~\cite{uavdt}}
        \label{fig:uavdt_me}
    \end{subfigure}

    \vspace{0.5em}

    \begin{subfigure}[t]{\linewidth}
        \centering
        \includegraphics[width=\linewidth]{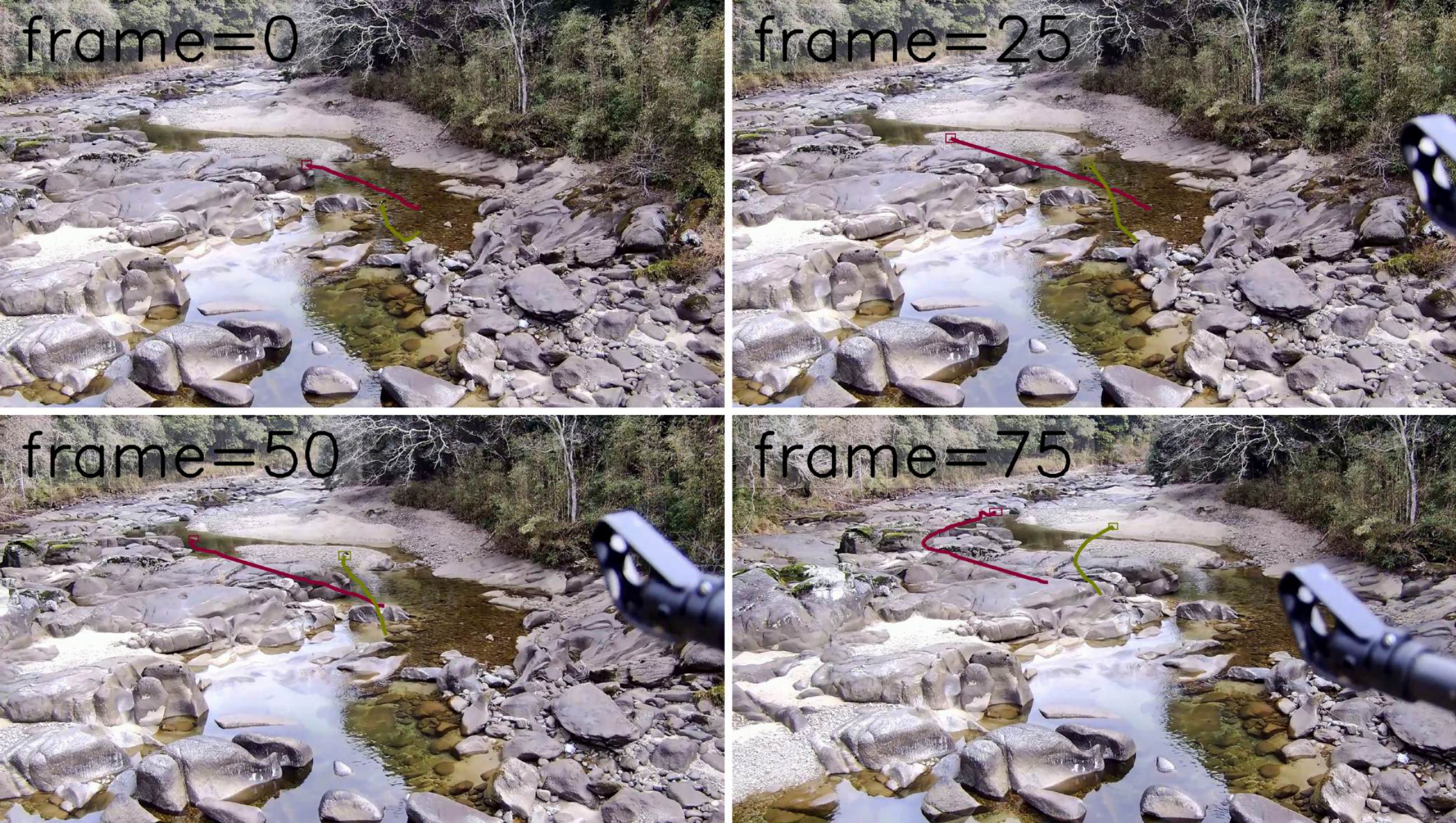}
        \caption{SMOT4SB dataset (Ours)}
        \label{fig:smot4sb_me}
    \end{subfigure}
    \vspace{-0.25em}
    \caption{Comparison of motion complexity across datasets. (a) MOT17 dataset assumes fixed cameras and ground-constrained pedestrian motion.
(b) UAVDT dataset captures vehicles with constrained planar motion from moving UAVs.
(c) SMOT4SB dataset features complex motion entanglement, where both the camera and subjects move independently in three-dimensional space, free from fixation or ground-contact constraints.}
    \label{fig:dataset_comparison}
\end{figure}

The unique challenges in SMOT4SB stem from several factors:

\noindent\textbf{Motion Entanglement:} As shown in Fig.~\ref{fig:dataset_comparison}, unlike conventional tracking problems with static cameras and planar object motion~\cite{campus,visdrone}, in SMOT4SB, both UAV platforms and objects move freely in three-dimensional space. This creates complex motion patterns that require sophisticated tracking strategies to disentangle ego-motion from target motion.

\noindent\textbf{Flocking Dynamics:} A group of birds move in coordinated patterns in many sequences.

\noindent\textbf{Scale Variation and Information Scarcity:} Consistent with findings from SOD4SB~\cite{mva2023_sod_challenge}, most bird instances occupy fewer than $32 \times 32$ pixels, leading to severe information scarcity that makes appearance-based association unreliable. Rapid object scale changes due to varying distances and flight patterns further exacerbate the detection consistency problem.


\section{SO-HOTA: Small Object Higher Order Tracking Accuracy}
\label{sec:so_hota}

\subsection{Motivation}
The widely adopted Intersection over Union (IoU) metric demonstrates excessive sensitivity to spatial displacement when evaluating small objects~\cite{xu2021dot}. This sensitivity poses fundamental challenges for small object tracking evaluation, where objects spanning merely a few pixels suffer from minimal positioning tolerance. When objects fall below $32 \times 32$ pixels, slight positional offsets that would be negligible for larger objects result in disproportionate performance penalties under IoU-based assessment~\cite{xu2021dot}.

The HOTA metric~\cite{luiten2021hota}, while addressing many limitations of earlier evaluation frameworks through its balanced DetA and AssA components, remains fundamentally dependent on IoU-based detection assessment. For small object scenarios, this dependency renders HOTA suboptimal because precise bounding box alignment becomes less critical than accurate center point localization. When considering small objects of $16 \times 16$ pixels, which frequently occur in SMOT scenarios, IoU-based evaluation presents two critical problems. Assuming predicted and ground-truth bounding boxes have identical sizes.

First, a minor displacement of just 8 pixels in the x-direction from perfect alignment causes IoU to drop dramatically from 1.00 to 0.5. This sharp decline triggers significant degradation in both AssA and DetA components of HOTA when evaluated against the standard threshold of 0.5 or higher, causing the evaluation 
\noindent metric to overreact to minor prediction errors and reducing the discriminative power for comparing different tracking methods.

Second, IoU-based evaluation fails to provide meaningful comparisons between different tracking results. Consider two tracking results: (1) non-overlapping predicted bounding boxes that are spatially close to ground-truth boxes, and (2) non-overlapping predicted bounding boxes that are spatially distant from ground-truth boxes. Both scenarios yield identical IoU values of zero, making it impossible to distinguish their relative quality in terms of AssA and DetA evaluation, despite clear differences in tracking accuracy.

The challenge becomes particularly acute in SMOT4SB scenarios involving \emph{motion entanglement}, where both camera platform and target objects exhibit complex three-dimensional motion patterns. The combination of small object sizes and nonlinear complex motion significantly increases the likelihood that predicted and ground-truth bounding boxes fail to overlap. This can cause the two aforementioned IoU-based evaluation problems to become prominent and potentially limit the effectiveness of tracking performance assessment.

\subsection{Integrating DotD with HOTA}

To address these limitations, the organizers propose SO-HOTA, which integrates the DotD measure with the HOTA evaluation metric.

\subsubsection{DotD Formulation}

DotD provides a normalized measure of center-point distance that exhibits reduced sensitivity to minor spatial displacements compared to IoU-based metrics.
 
Following the definition established by Xu et al.~\cite{xu2021dot}, DotD between two bounding boxes A and B is calculated as:

\begin{equation}
\text{DotD}(A,B) = e^{-\frac{D}{S}},
\end{equation}
where $D$ represents the Euclidean distance between bounding box centers:

\begin{equation}
D = \sqrt{(x_A - x_B)^2 + (y_A - y_B)^2},
\end{equation}
where $S$ denotes the average size of all objects in the dataset as follows:
\begin{equation}
S = \sqrt{\frac{\sum_{i=1}^{M}\sum_{j=1}^{N_i} w_{ij} \times h_{ij}}{\sum_{i=1}^{M} N_i}},
\end{equation}
where $M$ denotes the number of images, $N_i$ denotes the number of labeled bounding boxes in the $i$-th image, and $w_{ij}$, $h_{ij}$ represent the width and height of the $j$-th bounding box in the $i$-th image.

\subsubsection{SO-HOTA Metric Definition}

HOTA is consists of DetA and AssA.
Iou in DetA is replaced by DotD in SO-DetA, which is used in our SO-HOTA, as follows:
\begin{equation}
\text{SO-DetA}_{\alpha} = \frac{|\text{TP}_{\text{DotD}}|}{|\text{TP}_{\text{DotD}}| + |\text{FN}| + |\text{FP}|},
\end{equation}
where $\text{TP}_{\text{DotD}}$ represents true positives determined using a DotD threshold, $\alpha$, instead of an IoU threshold. A detection pair is considered a true positive when $\text{DotD}(A,B) \geq \alpha$.

As with DetA mentioned above, IoU in AssA is also replaced by DotD in our SO-AssA as follows:
\begin{equation}
\text{SO-AssA}_{\alpha} = \frac{1}{|\text{TP}_{\text{DotD}}|} \sum_{c \in \{\text{TP}_{\text{DotD}}\}} A(c),
\end{equation}
where $A(c)$ denotes the association accuracy for true positive $c$:
\begin{equation}
A(c) = \frac{|\text{TPA}(c)|}{|\text{TPA}(c)| + |\text{FNA}(c)| + |\text{FPA}(c)|}
\end{equation}

The SO-HOTA score integrates detection and association accuracy through geometric mean:
\begin{equation}
\text{SO-HOTA}_{\alpha} = \sqrt{\text{SO-DetA}_{\alpha} \times \text{SO-AssA}_{\alpha}}
\end{equation}

Following HOTA, the comprehensive SO-HOTA score averages across multiple DotD thresholds:
\begin{equation}
\text{SO-HOTA} = \frac{1}{19} \sum_{\alpha \in \{0.05, 0.1, \ldots, 0.95\}} \text{SO-HOTA}_{\alpha}
\end{equation}

SO-HOTA follows the same matching procedure as HOTA, using the Hungarian algorithm to assign predicted and ground-truth detections in a one-to-one manner. DotD is used as the similarity measure instead of IoU, allowing better tolerance to slight localization errors common in small object tracking. The matching prioritizes high similarity pairs that meet a DotD threshold, while preserving HOTA’s optimization structure for detection, association, and localization accuracy.

\begin{figure*}[t]
  \begin{center}
    \includegraphics[width=\linewidth]{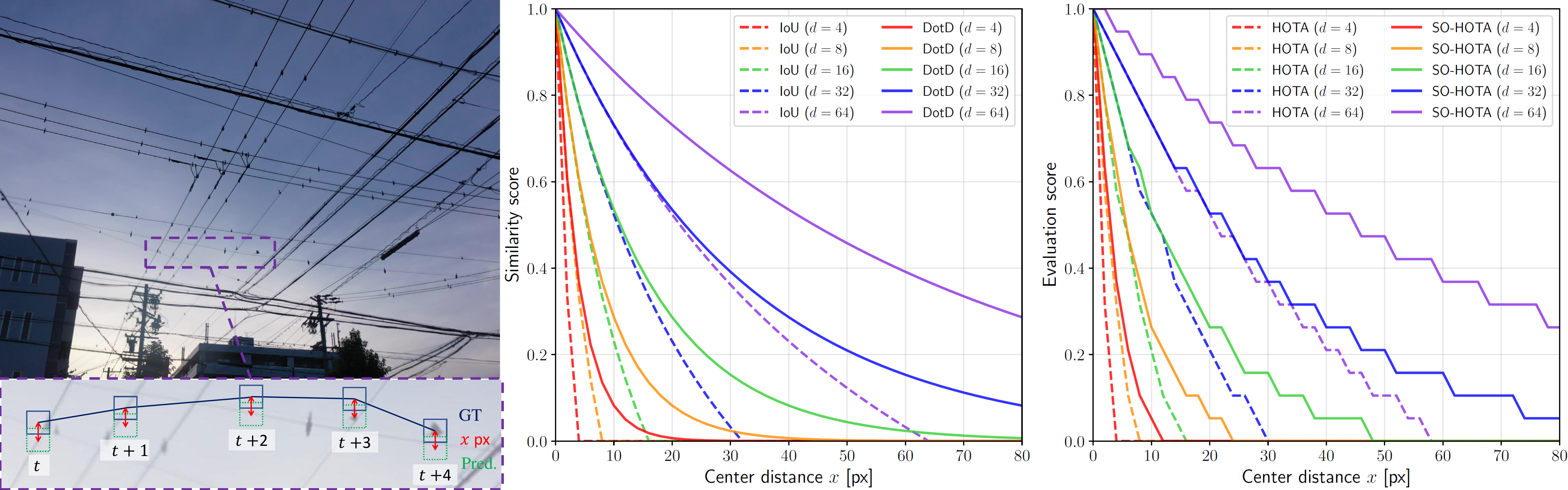}
  \end{center}
  \vspace{-3mm}
        \caption{Evaluation of HOTA and SO-HOTA under synthetic displacement.
(Left) Tracking scenario where predicted boxes are vertically shifted by $x$ pixels from ground truth.
(Center) IoU and Dot Distance (DotD) scores versus center distance $x$ for various box sizes $d$.
(Right) HOTA changes sensitively with even 1-pixel displacement and drops to zero once overlap is lost ($x > d/2$). In contrast, SO-HOTA decays smoothly based on spatial proximity, enabling more stable evaluation.}
    \label{fig:evaluation_comparison}
\end{figure*}

\subsection{Analysis of Evaluation Metrics}

To demonstrate the limitations of HOTA and the advantages of SO-HOTA for small object tracking evaluation, we present a comparative analysis based on center displacement scenarios, as illustrated in Fig.~\ref{fig:evaluation_comparison}.

The left graph of Fig.~\ref{fig:evaluation_comparison} visualizes a tracking scenario where predicted bounding boxes are consistently displaced vertically by $x$ pixels from the ground truth across frames, simulating typical localization errors in small object tracking. The center panel compares the similarity scores of IoU and Dot Distance (DotD) as a function of the center distance $x$, under various bounding box sizes $d$. As shown, IoU drops sharply to zero when $x$ exceeds $d/2$, making it highly sensitive to small displacements. In contrast, DotD decays smoothly and continuously, allowing finer discrimination based on spatial proximity.

The right graph of Fig.~\ref{fig:evaluation_comparison} compares the resulting HOTA and SO-HOTA scores under the same displacement conditions. Due to its reliance on IoU, HOTA exhibits steep score drops and fails to distinguish moderate displacements, especially for small $d$. Once the center displacement exceeds $d/2$, the predicted and ground-truth boxes no longer overlap, causing IoU to become zero and HOTA to drop to zero accordingly. Furthermore, depending on the bounding box size, particularly when $d = 32$ or $64$, HOTA can become zero even when IoU is still non-zero. In contrast, SO-HOTA leverages DotD for matching and scoring, and continues to assign non-zero evaluation scores based on spatial proximity, even without bounding box overlap. This enables more stable and meaningful comparisons in small object tracking scenarios where minor misalignments are common.


\section{MVA 2025 Small Multi-Object Tracking for Spotting Birds
Challenge}
\label{sec:challenge}

We describe the details of the challenge using the SMOT4SB dataset.

\subsection{Baseline Code}

The organizers provide a baseline implementation for the SMOT4SB challenge, which consists of a two-stage tracking-by-detection pipeline. The first stage employs a YOLOX detector~\cite{yolox2021} fine-tuned on the SMOT4SB dataset to localize small flying objects. The second stage applies OC-SORT~\cite{cao2023observation} to generate trajectory-level predictions. This baseline provides a solid foundation for participants. Please refer to Appendix~\ref{sec:apd:resource} for the code repository link.

\subsection{Challenge Phases}

The public and private test phases were given to participants.
During the public test phase, participants can evaluate their methods on the public test subset of the SMOT4SB dataset by submitting their detection results to CodaBench~\cite{codabench}.
In this phase, the participants can access only images without annotations.
In the private test phase, the organizer ran the code provided by each team for evaluation on the private test subset of the SMOT4SB dataset.
After the challenge ends, CodaBench is publicly available for evaluation on the public test subset, as described in Appendix~\ref {sec:apd:resource}.
Each team can evaluate their results at most 10 times per day for restricting HARKing~\cite{kerr1998harking} to the public test subset.


\section{Challenge Results}

\begin{table*}[t!]
  \centering
  \caption{Quantitative results from the \textbf{private test} for the top 5 teams in this challenge (ranked by SO-HOTA).}
  \label{tab:pri_test_result}
  \resizebox{\linewidth}{!}{
  \begin{tabular}{c|c|rrrrr|rrrrr}
    \toprule
    \multirow{2.5}{*}{Rank} & \multirow{2.5}{*}{Team} & \multicolumn{5}{c|}{\textbf{SO-HOTA suite}} & \multicolumn{5}{c}{Previous metrics} \\
    \cmidrule(lr){3-7} \cmidrule(lr){8-12}
    & & \textbf{SO-HOTA}$\uparrow$ & SO-DetA$\uparrow$ & SO-AssA$\uparrow$ & SO-DetRe$\uparrow$ & SO-DetPr$\uparrow$ & HOTA$\uparrow$ & MOTA$\uparrow$ & IDF1$\uparrow$ & MT$\uparrow$ & ML$\downarrow$ \\
    \midrule
    1 & DL Team~\cite{yu_yolov8-smot_2025} & \textbf{50.59} & 47.27 & 54.30 & 52.67 & 80.38 & 36.74 & 28.80 & 44.86 & 0.26 & 0.37 \\
    2 & zhwa2003~\cite{zhwa2003_smot4sb} & \textbf{46.22} & 42.69 & 50.28 & 48.71 & 75.35 & 31.82 & 15.87 & 37.61 & 0.19 & 0.45 \\
    3 & elsalabA~\cite{elsalabA_smot4sb}& \textbf{43.87} & 40.13 & 48.11 & 53.39 & 59.96 & 28.60 & -8.12 & 33.80 & 0.09 & 0.47 \\
    4 & sgm~\cite{sgm_smot4sb} & \textbf{43.71} & 43.85 & 43.72 & 49.97 & 76.57 & 32.81 & 28.15 & 38.68 & 0.24 & 0.37 \\
    5 & xmba15 & \textbf{40.49} & 31.85 & 51.59 & 46.46 & 49.51 & 28.94 & -8.11 & 35.12 & 0.17 & 0.40 \\
    \midrule
    - & Baseline & \textbf{9.90} & 8.67 & 11.32 & 8.83 & 80.67 & 6.51 & 1.61 & 5.18 & 0.00 & 0.91 \\
    \bottomrule
  \end{tabular}
  }
\end{table*}

\begin{table*}[t!]
  \centering
  \caption{Quantitative results from the \textbf{public test} for the top 5 teams (ranked by SO-HOTA).}
  \label{tab:pub_test_result}
  \resizebox{\linewidth}{!}{
  \begin{tabular}{c|c|rrrrr|rrrrr}
    \toprule
    \multirow{2.5}{*}{Rank} & \multirow{2.5}{*}{Team} & \multicolumn{5}{c|}{\textbf{SO-HOTA suite}} & \multicolumn{5}{c}{Previous metrics} \\
    \cmidrule(lr){3-7} \cmidrule(lr){8-12}
    & & \textbf{SO-HOTA}$\uparrow$ & SO-DetA$\uparrow$ & SO-AssA$\uparrow$ & SO-DetRe$\uparrow$ & SO-DetPr$\uparrow$ & HOTA$\uparrow$ & MOTA$\uparrow$ & IDF1$\uparrow$ & MT$\uparrow$ & ML$\downarrow$ \\
    \midrule
    1 & DL Team~\cite{yu_yolov8-smot_2025}   & \textbf{54.90} & 51.22 & 59.02 & 58.65 & 78.55 & 41.67 & 36.86 & 51.94 & 0.34 & 0.22 \\
    2 & sgm~\cite{sgm_smot4sb}       & \textbf{53.85} & 51.31 & 56.60 & 59.97 & 76.69 & 41.38 & 37.15 & 50.74 & 0.18 & 0.43 \\
    3 & zhwa2003~\cite{zhwa2003_smot4sb}  & \textbf{49.20} & 44.03 & 55.15 & 51.92 & 72.73 & 36.05 & 24.16 & 44.25 & 0.16 & 0.30 \\
    4 & elsalabA~\cite{elsalabA_smot4sb}  & \textbf{44.54} & 34.45 & 57.65 & 43.12 & 61.19 & 30.28 & -0.07 & 35.76 & 0.06 & 0.59 \\
    5 & xmba15    & \textbf{40.92} & 34.02 & 49.31 & 56.48 & 45.41 & 30.06 & -20.78 & 36.08 & 0.13 & 0.35 \\
    \midrule
    - & Baseline & \textbf{10.68} & 9.79 & 11.67 & 10.07 & 75.56 & 6.74 & 0.00 & 5.83 & 0.00 & 0.94 \\
    \bottomrule
  \end{tabular}
  }
\end{table*}

\begin{table}[t!]
  \centering
  \caption{Comparison of model performance. $^\dag$ and $^\ddag$ indicate the GPUs used for training and inference, respectively.}
  \label{tab:model_performance}
  \resizebox{\linewidth}{!}{
  \begin{tabular}{c|rr|l}
    \toprule
    \multirow{2.5}{*}{Team} &
    \multicolumn{2}{c|}{Resource usage} &
    \multirow{2.5}{*}{GPU} \\
    \cmidrule(lr){2-3}
    & Params. [M] & Time [s/img] & \\
    \midrule
    DL Team~\cite{yu_yolov8-smot_2025}   & 44  & 0.175 & RTX3090 \\
    zhwa2003~\cite{zhwa2003_smot4sb}  & 215 & 0.111 & RTX3060 \\
    elsalabA~\cite{elsalabA_smot4sb}  & 304    & 0.500 & RTX3090 \\
    sgm~\cite{sgm_smot4sb}       & 57   & 0.107 & RTX4090$^\dag$, RTXA6000$^\ddag$ \\
    xmba15    &  64 & 0.434 & GTX1080Ti \\
    \midrule
    Baseline &  99 & 0.189 & GTX1080Ti \\
    \bottomrule
  \end{tabular}
  }
\end{table}

\subsection{Participation Overview}

78 participants joined this challenge, resulting in a total of 308 submissions. Finally, 7 teams provided their final solutions for evaluation on the private test set.
The primary ranking criterion for this challenge is the SO-HOTA score, with the teams ranked based on their highest achieved score. In addition, for reference, conventional metrics such as HOTA~\cite{luiten2021hota}, MOTA~\cite{bernardin2008evaluating}, IDF1~\cite{idf1}, Mostly Tracked targets (MT), and Mostly Lost targets (ML) will also be evaluated.

\subsection{Performance Results and Analysis}
Tables~\ref{tab:pri_test_result} and~\ref{tab:pub_test_result} present the quantitative results of the top 5 teams in the SMOT4SB challenge. The winning team, DL Team, achieved a SO-HOTA score of 50.59 on the private test set, a remarkable 5.1$\times$ improvement over the baseline score of 9.90. This substantial performance gain demonstrates the effectiveness of specialized approaches for small multi-object tracking.
The results reveal a notable performance gap between SO-HOTA and traditional HOTA scores across all methods. For instance, DL Team achieved 50.59 in SO-HOTA but only 36.74 in HOTA.

\begin{figure}[t]
  \centering
  \includegraphics[width=\linewidth]{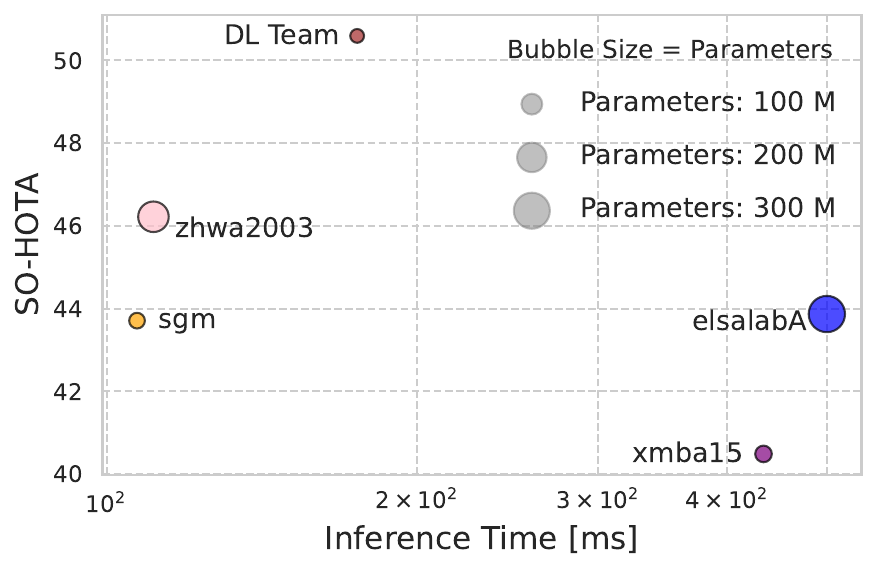}
  \caption{
      SO-HOTA vs. inference time for the winning methods on the SMOT4SB private test subset. The horizontal axis (in seconds per image) is shown in log scale. Marker size indicates the number of model parameters.
    }
  \vspace{-0.8em}
  \label{fig:params_score}
\end{figure}

Table~\ref{tab:model_performance} and Fig.~\ref{fig:params_score} jointly highlight the diverse design trade-offs among the top-performing methods. DL Team achieved the highest SO-HOTA score (54.90) with a compact model (44M parameters) and low inference cost (0.175 s/img), representing a well-optimized solution in both accuracy and efficiency. On the other hand, sgm offered a different trade-off, combining the fastest inference (0.107 s/img) and a moderate model size (57M) with a slightly lower but still competitive SO-HOTA of 43.71. This comparison demonstrates varied strategies in balancing accuracy, model complexity, and efficiency.

\section{Challenge Methods and Teams}

This section provides a brief description of the methods proposed by the winning teams. 

\subsection{DL Team}

The winning solution from DL Team is an efficient tracking-by-detection framework~\cite{sort_2016}, composed of a specialized small object detector, \textbf{YOLOv8-SOD}, and a robust, appearance-free tracker, \textbf{YOLOv8-SMOT}~\cite{yu_yolov8-smot_2025}. Notably, as shown in Table~\ref{tab:model_performance}, our method achieved the highest score while utilizing the fewest model parameters among all top-5 teams. The core of our method is a novel training strategy named \textbf{SliceTrain}, designed to significantly boost detection performance on minuscule targets without incurring inference-time overhead.

\begin{figure}[t]
 \centering
 \includegraphics[width=\linewidth]{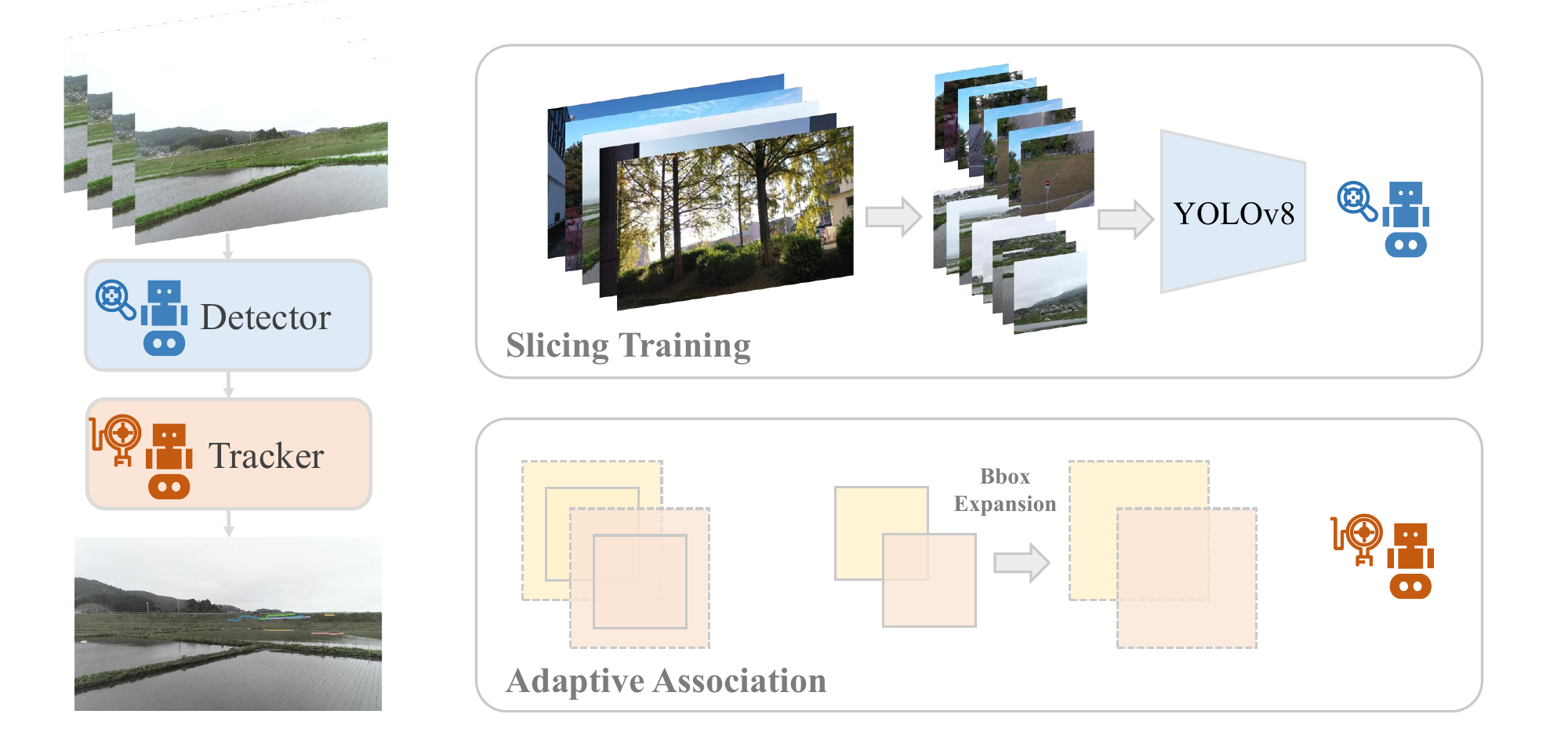}
 \caption{
      Overview of the DL Team's method. It showcases the core \textbf{SliceTrain} strategy (slicing for training, full-image for inference) and the key enhancements in the \textbf{YOLOv8-SMOT}~\cite{yu_yolov8-smot_2025} tracker (Motion Direction Maintenance and Adaptive Similarity Metric).
     }
 \label{fig:dl_team_overview}
\end{figure}

Our detector, YOLOv8-SOD, is based on the powerful YOLOv8 architecture~\cite{Jocher_Ultralytics_YOLO_2023}. It is fine-tuned using the \textbf{SliceTrain} strategy, where high-resolution training images are partitioned into smaller, overlapping patches. This technique allows for a substantially larger training batch size on standard hardware, enabling the model to learn more discriminative features for small objects. Critically, during inference, the fully-trained model is applied directly to the original, full-resolution images. This asymmetric "slice for training, full-image for inference" approach avoids the computational cost and potential stitching errors of methods that require slicing at inference time.

Our tracker, YOLOv8-SMOT, enhances the OC-SORT framework~\cite{cao2023observation} to handle the erratic motion and small size of birds. It introduces two key improvements: (1) \textbf{Motion Direction Maintenance}, which uses an Exponential Moving Average (EMA) on velocity vectors to create a more stable motion direction cue for association. (2) An \textbf{Adaptive Similarity Metric} that replaces the standard IoU by combining bounding box expansion with a distance penalty, which is similar to \cite{huang_iterative_2024,yang_hard_2023}. This provides a reliable association signal even when small targets do not have overlapping bounding boxes between frames. The effectiveness of these tracker enhancements is demonstrated in the ablation study in Table~\ref{tab:dl_team_ablation}.

\begin{table}[t]
 \centering
  \caption{Ablation study of our tracker components on the public test set with the YOLOv8-L detector. `Default` refers to the OC-SORT baseline.}
 \label{tab:dl_team_ablation}
 \resizebox{0.9\linewidth}{!}{
 \begin{tabular}{lccc}
   \toprule
   \textbf{Method} & \textbf{SO-HOTA}$\uparrow$ & \textbf{SO-DetA}$\uparrow$ & \textbf{SO-AssA}$\uparrow$ \\
   \midrule
   Default (OC-SORT) & 44.20 & 46.09 & 42.53 \\
   + EMA & 47.92 & 48.02 & 48.01 \\
   + Bbox Expansion & 51.39 & 51.43 & 51.49 \\
   + Distance Penalty (\textbf{YOLOv8-SMOT}) & \textbf{55.21} & \textbf{51.72} & \textbf{59.08} \\
   \bottomrule
 \end{tabular}
  }
\end{table}

The performance and efficiency trade-offs of our framework are presented in Table~\ref{tab:dl_team_perf_speed}. While the largest model (YOLOv8-L) achieves the highest accuracy, our smaller models offer compelling alternatives for resource-constrained scenarios. The YOLOv8-S model, for instance, reaches a real-time speed of \textbf{17.61 FPS} with only a minor drop in the SO-HOTA score, demonstrating its suitability for practical, real-world applications.

Furthermore, the efficiency of our models can be significantly enhanced through model quantization. By applying advanced Quantization-Aware Training (QAT) methods like GPLQ~\cite{liang2025gplq} or practical Post-Training Quantization (PTQ) techniques like QwT~\cite{fu2025quantization} and QwT-v2~\cite{tang2025qwt}, the computational footprint can be further reduced. This would enable our high-performance tracking framework to be deployed on low-power edge devices while maintaining real-time capabilities.

\begin{table}[t]
	\centering
	\caption{Performance vs. Efficiency on the public test set. Memory and Speed were measured on a single NVIDIA RTX 3090 GPU with an image resolution of $2160\times 3840$.}
	\label{tab:dl_team_perf_speed}
	\resizebox{\linewidth}{!}{
	\begin{tabular}{lccccc}
		\toprule 
		Model & SO-HOTA & SO-DetA & SO-AssA & Param.(M) & Speed (FPS) \\
		\midrule 
		YOLOv8-L & \textbf{55.21} & \textbf{51.72} & 59.08 &  43.7 & 5.70 \\
		YOLOv8-M & 54.43 & 49.53 & 59.96 &  25.9 & 8.96 \\
		YOLOv8-S & 53.81 & 48.39 & 59.98 & 11.2  & \textbf{17.61} \\
		\bottomrule
	\end{tabular}
	}
\end{table}

\subsection{zhwa2003}
Zhwa2003 team proposed an intersection-based ensemble method~\cite{zhwa2003_smot4sb} that effectively filters out false positives in specific challenging scenes. They constructed a robust base model (RB model) which uses Cascade R-CNN~\cite{cai_cascade_2018} as the model architecture, with Swin Transformer~\cite{liu_swin_2021} as the backbone for visual feature extraction, and Gaussian Receptive Field based Label Assignment (RFLA)~\cite{xu_rfla_2022} for loss calculation. To deal with the challenging scenes, especially with power poles in urban areas, they constructed a refined model (RS model) on the custom PPSM dataset with the same architecture as the RB model. Around 66\% of the images in the PPSM dataset contain power poles. 

\begin{figure}[t]
  \centering
  \includegraphics[width=\linewidth]{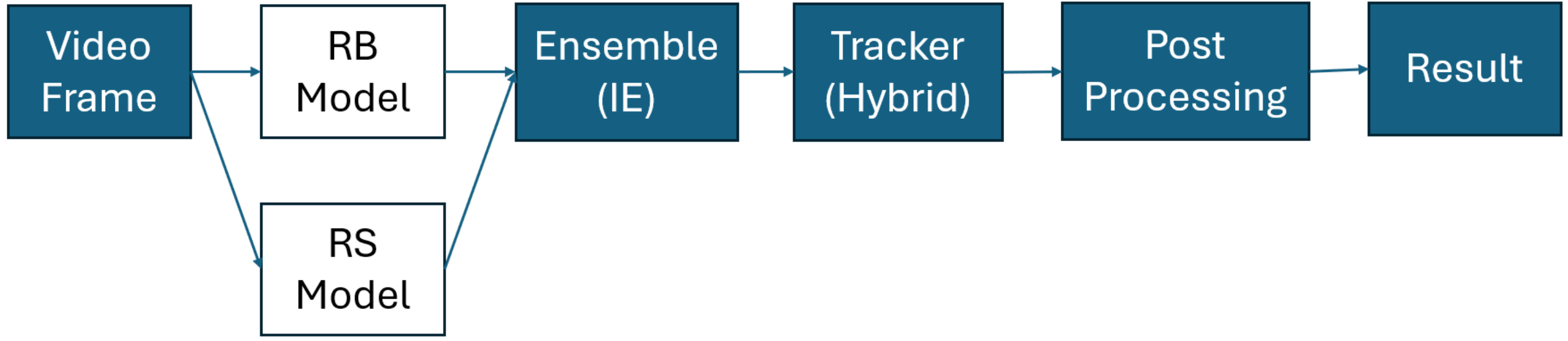}
  \caption{
       Overview of the proposed system. “IE” means intersection-based ensemble strategy. “Hybrid” is Hybrid-SORT.
    }
  \label{fig:zhwa_overview}
\end{figure}

Fig.~\ref{fig:zhwa_overview} shows an overview of the system. The interaction-based ensemble strategy takes results from both the RB and RS models as input. For each pair of bounding boxes from the two models, they calculate their IoU. If the IoU between a pair of boxes is greater than zero, they select the detection result from the RB model. If IoU is zero, they ignore this pair in the following processes. By requiring both models to agree on the presence of an object via intersection, and by choosing the RB model’s prediction, they effectively filter out false positives from the RB model that are not confirmed by the RS model. This process provides more accurate detection results for the tracker Hybrid-SORT~\cite{yang_hybrid-sort_2024}. In the end, they apply interpolation to the tracker’s output.

\begin{figure}[t]
  \centering
  \includegraphics[width=\linewidth]{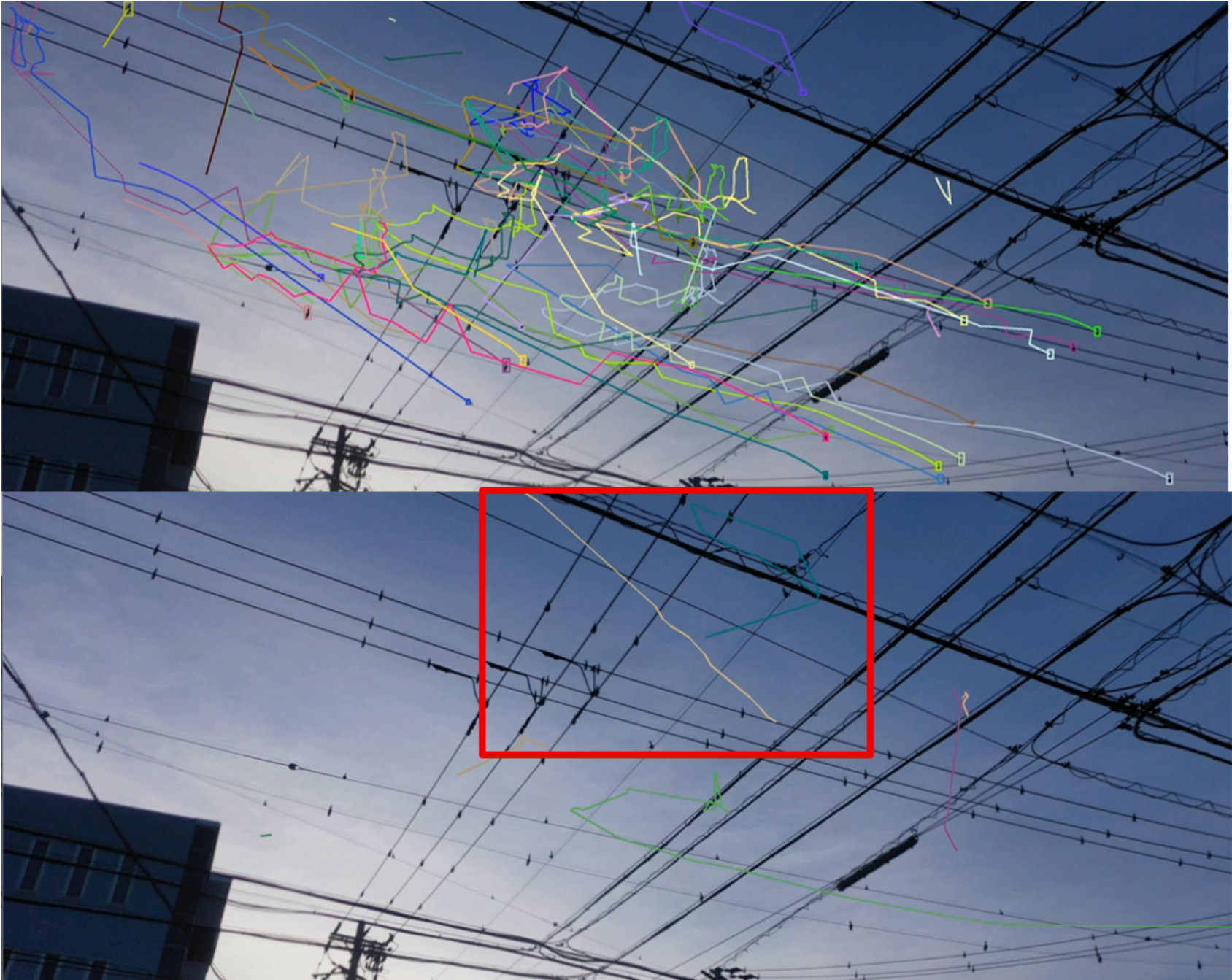}
  \caption{
       The upper image shows the RB model’s result. The lower image shows the result obtained by the IE strategy, where only the track in the red box is true.
    }
  \label{fig:zhwa_visualize_IE}
\end{figure}

Table~\ref{tab:zhwa_IE_tab} shows the effectiveness of the intersection-based ensemble strategy on the public test dataset of the MVA2025 Challenge. The Small Object Detection Precision (SO-DetPr) of the tracking results is increased by 20.6 points compared to the RB-only method. A substantial decrease in false positives can be observed while maintaining an almost identical number of true positives. This precise filtering directly contributes to an increase of 4.17 in SO-HOTA. Fig.~\ref{fig:zhwa_visualize_IE} shows the tracking results without and with the intersection-based ensemble strategy.

\begin{table}[t]
  \centering
  \caption{Performance of the RB-only method and that with the IE strategy.}
  \label{tab:zhwa_IE_tab}
  \resizebox{\linewidth}{!}{
  \begin{tabular}{lccccc}
    \toprule
    Method & SO-HOTA & SO-DetA & SO-DetRe & SO-DetPr & CLR\_FP \\
    \midrule
    RB only        & 42.96 & 35.54 & 47.77 & 57.29 & 19,091 \\
    RB+RS with IE  & 47.13 & 42.28 & 47.47 & 77.88 & 7,629 \\
    \bottomrule
  \end{tabular}
  }
\end{table}

\subsection{elsalabA}

\begin{figure}[t]
    \centering
    \includegraphics[width=0.8\linewidth]{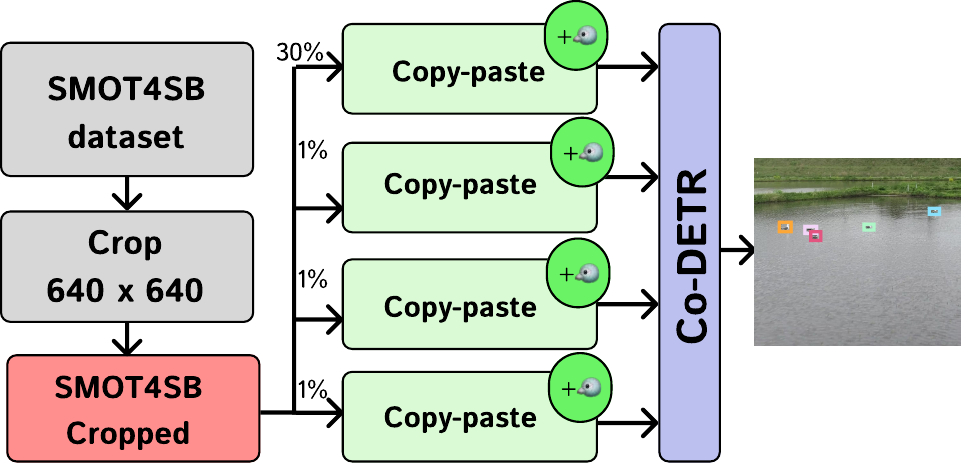}
    \caption{Team elsalabA Training Method Overview.}
    \label{fig:elsalabA_training}
    \vspace{-0.5em}
\end{figure}

\begin{figure}[t]
    \centering
    \includegraphics[width=0.8\linewidth]{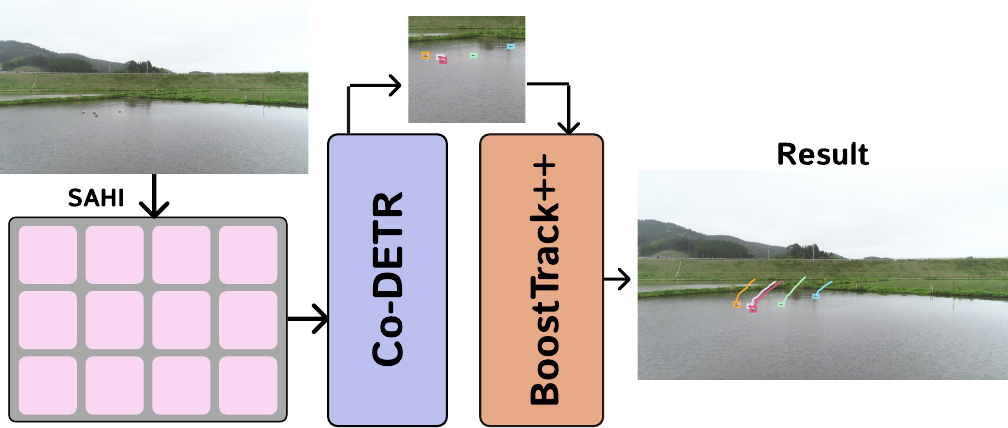}
    \caption{Team elsalabA Inference Procedure Overview}
    \label{fig:elsalabA_inference}
\end{figure}

Team elsalabA designed a two-stage detection and tracking pipeline~\cite{elsalabA_smot4sb} that combines Collaborative Hybrid Assignments DETR (Co-DETR)~\cite{zong2023detrs} with the multi-object tracker BoostTrack++~\cite{stanojevic2024boostTrack,stanojevic2024btpp}. To address the challenge of detecting extremely small objects in aerial footage, elsalabA further incorporated small object detection techniques, including Copy-Paste augmentation~\cite{copy_paste_augmentation} and Slicing Aided Hyper Inference (SAHI)~\cite{akyon2022sahi, obss2021sahi}. Their complete pipeline improved performance on the SMOT4SB dataset, achieving a SO-HOTA score of 43.87 on the private test set, as summarized in Table~\ref{tab:pri_test_result}.

During training, the original high-resolution SMOT4SB images were cropped to a resolution of $640\times640$. To increase small object diversity and the number of positive samples, elsalabA applied an iterative Copy-Paste augmentation strategy. As illustrated in Fig.~\ref{fig:elsalabA_training}, bird instances from the Birds Flying Dataset~\cite{birdsflyingdataset} and synthetic birds generated using Stable Diffusion~\cite{rombach2022high} were composited onto diverse backgrounds over the stages. This gradually increased the visual complexity and object density of training samples, enabling the Co-DETR detector to learn more robust representations of small objects.

 At inference time, elsalabA employed SAHI to tile high-resolution test images into overlapping crops, consistent with the training resolution (see Fig.~\ref{fig:elsalabA_inference}). This not only enhanced the detection of small objects but also allowed inference under limited GPU memory. Each tile was processed independently by the fine-tuned Co-DETR model, and the individual detections were aggregated to form a complete scene-level prediction. The final detections were then passed to BoostTrack++, which maintained temporal consistency and preserved object identities across frames.

As shown in Table~\ref{tab:elsalabA_ap50} and \ref{tab:elsalabA_ablation}, elsalabA conducted ablation studies to evaluate each component's contribution. The removal of Copy-Paste or replacement of BoostTrack++ with another tracker caused the drop in detection and tracing metrics, respectively, validating the effectiveness of both techniques. 

\begin{table}[t!]
\renewcommand{\arraystretch}{1.0} 
\caption{Detection ablation study on cropped validation set (640$\times$640). The second row (i.e., \textit{without paste}) removed the Copy-Paste augmentation during training. The third row (i.e., \textit{without SD}), excluded the synthetic instances generated by Stable Diffusion during training stage, and only pasted the birds from Birds Flying Dataset to the background.}
\label{tab:elsalabA_ap50}
\renewcommand{\arraystretch}{1.0}
\footnotesize
\centering
\begin{tabular}{l p{1.7cm}<{\centering} p{1.7cm}<{\centering} p{1.7cm}<{\centering} p{1.5cm}<{\centering}}
\toprule
\textbf{Detection Training} & \textbf{AP@0.50} & \textbf{AP@0.75} \\
\midrule
Co-DETR & \textbf{0.59} & \textbf{0.13} \\
Co-DETR w/o paste & 0.54 & 0.11 \\
Co-DETR w/o SD & 0.55 & 0.12 \\
\bottomrule
\end{tabular}
\end{table}

\begin{table}[t!]
\renewcommand{\arraystretch}{1.0}
\caption{Tracking ablation study on the public dataset. On the second row (i.e., \textit{with BoT-SORT}), elsalabA replaced BoostTrack++ with BoT-SORT during tracking.}
\label{tab:elsalabA_ablation}
\renewcommand{\arraystretch}{1.0}
\footnotesize
\centering
\resizebox{\linewidth}{!}{
\begin{tabular}{l p{1.7cm}<{\centering} p{1.7cm}<{\centering} p{1.7cm}<{\centering} p{1.5cm}<{\centering}}
\toprule
\textbf{Method} & \textbf{SO-HOTA} & \textbf{SO-DetRe} & \textbf{SO-DetPr} \\
\midrule
elsalabA Pipeline & \textbf{42.00}    & \textbf{50.65}  & \textbf{47.18} \\
w/ BoT-SORT~\cite{aharon2022bot}      & 8.68   & 16.70  & 40.01 \\
\bottomrule
\end{tabular}
}
\vspace{-0.5em}
\end{table}

\subsection{sgm}
The sgm team proposed a framework that combines ensemble detection from multiple YOLO models with Adaptive Weighted Boxes Fusion, which dynamically adjusts fusion weights based on confidence scores rather than static performance metrics~\cite{sgm_smot4sb}.

In the proposed method, YOLOv7~\cite{wang2023yolov7} and YOLOv12~\cite{tian2025yolov12} are used as complementary object detectors. YOLOv7 is applied at multiple high resolutions (3200, 3360, and 3560 pixels), and an additional output is obtained by applying horizontal flip augmentation at 3200 resolution, resulting in a total of four outputs. YOLOv12 is applied at 1280 resolution with horizontal flip augmentation, yielding two outputs. This multi-scale and flip-based inference is expected to enhance detection performance for objects of varying sizes.

Unlike conventional fixed-weight WBF~\cite{solovyev2021wbf} or methods that rely on precomputed performance metrics, the proposed Adaptive WBF dynamically adjusts weights in each frame based on the average confidence scores of detections as follows:

\begin{equation}
\bar{s_i} = \frac{1}{n_i} \sum_{j=1}^{n_i} s_{i,j}, \quad w_i = \frac{\bar{s_i}}{\sum_k \bar{s_k}}
\end{equation}

where $n_i$ is the number of detections from detector $i$, and $s_{i,j}$ is the confidence score of the $j$-th detection. This approach allows the system to assign higher weights to more reliable detectors in each frame, which is expected to improve detection stability under fluctuating conditions.

For tracking, we employ OC-SORT~\cite{cao2023observation}, using Distance-IoU~\cite{zheng2020distance} instead of conventional IoU to handle cases where bounding boxes do not overlap between consecutive frames due to rapid bird motion or camera movement. This modification is expected to enhance tracking robustness in low-overlap situations commonly observed in UAV footage.

\subsection{xmba15}

Team xmba15 proposed a tracking-by-detection framework, incorporating a detector extended from the SOD4SB~\cite{mva2023_sod_challenge} challenge baseline and motion-compensated tracking. For object detection, we employed an ensemble of three CenterNet~\cite{zhou2019objects} models trained on the SMOT4SB dataset and a combination of the SOD4SB and FBD-SV-2024~\cite{sun2025fbd} datasets, the latter comprising surveillance videos of flying birds in cluttered environments. One model was trained solely on the SMOT4SB dataset using an EfficientNet-B1~\cite{tan2019efficientnet} backbone, while the other two, using EfficientNet-B1 and RexNet-150~\cite{han2021rethinking} backbones, were trained on the merged dataset. Training was performed on $640\times640$ image patches, while inference was conducted on full-resolution frames. The outputs of all three detectors were fused using Weighted Box Fusion (WBF)~\cite{solovyev2021wbf} to enhance detection accuracy. To further reduce false positives, we trained a binary bird classifier using ground truth crops and false positives from the ensemble output, and applied it to filter the final detections.

\begin{figure}[t]
    \centering
    \includegraphics[width=\linewidth]{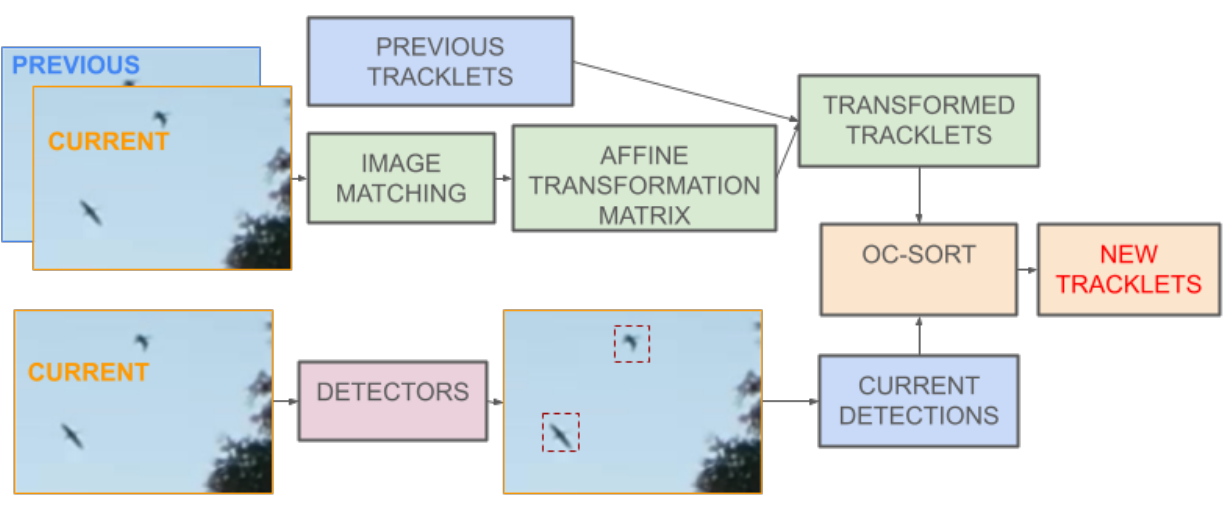}
    \caption{Team xmba15 Framework overview.}
    \label{fig:xmba15_framework}
    \vspace{-1em}
\end{figure}

For tracking, we adopted the OC-SORT~\cite{cao2023observation} algorithm, as used in the SMOT4SB baseline. However, we found its performance degraded under rapid ego-motion of the camera. To mitigate this, we introduced an image matching pipeline that estimates the affine transformation between consecutive frames. We extracted keypoints using the DISK~\cite{tyszkiewicz2020disk} detector and performed matching with LightGlue~\cite{lindenberger2023lightglue}. The estimated affine matrix was then used to update the positions of bounding boxes from previous tracklets. OC-SORT subsequently associated current-frame detections with these motion-compensated tracklets, improving tracking stability under dynamic camera conditions.

The overall structure of the framework is illustrated in Fig.~\ref{fig:xmba15_framework}.

\section{Conclusion}
\label{sec:conclusion}

This paper presents the SMOT4SB challenge, advancing small multi-object tracking through a comprehensive dataset, novel evaluation methodology, and algorithmic innovation. The SMOT4SB dataset provides the first large-scale collection for small object tracking in UAV scenarios with motion entanglement. The proposed SO-HOTA metric effectively addresses the limitations of IoU-based evaluation for SMOT. The challenge drove significant algorithmic advancement, with the winning method achieving 5.1× improvement over the baseline.

Future work includes extending to trajectory forecasting tasks, introducing confidence-aware tracking evaluation, and developing drone-based applications for bird strike avoidance, agricultural and fisheries protection, and ecological monitoring through automated population surveys.

\section{Acknowledgments}
We would like to thank Dr. Masatsugu Kidode at Nara Institute of Science and Technology for his helpful advice. We also thank ProDrone Co., Ltd. for serving as our technical partner in capturing images using UAVs. We also appreciate code testers and annotators.

A donation from the MVA organization supported this challenge.

\bibliographystyle{ieeetr}
\bibliography{reference.bib}

\appendix
\section{Teams and Affiliations}
\label{sec:apd:team}

\subsection{MVA 2025 SMOT4SB Challenge Organizers}
\noindent{\textbf{Title: }}\\ MVA 2025 Small Multi-Object Tracking for Spotting Birds Challenge: Dataset, Methods, and Results\\

\noindent{\textbf{Members:}}\\ \textbf{Yuki Kondo$^1$ (yuki\_kondo\_ab@mail.toyota.co.jp)}, Norimichi Ukita$^2$ , Riku Kanayama$^2$, Yuki Yoshida$^2$, Takayuki Yamaguchi$^3$\\

\noindent{\textbf{Affiliations: }}\\
$^1$ Toyota Motor Corporation, Japan\\
$^2$ Toyota Technological Institute, Japan\\
$^3$ Iwate Prefecture Coastal Regional Development Bureau, Japan\\

\subsection{DL Team}
\noindent{\textbf{Title:}}\\ YOLOv8-SMOT: An Efficient and Robust Framework for Real-Time Small Object Tracking via Slice-Assisted Training and Adaptive Association  \\

\noindent{\textbf{Members:}}\\
\textbf{Xiang Yu$^1$ (221300049@smail.nju.edu.cn)}, Guang Liang$^1$, Xinyao Liu$^2$ \\

\noindent{\textbf{Affiliations:}}\\
$^1$ Nanjing University, China\\
$^2$ University of Science and Technology of China, China

\subsection{zhwa2003}
\noindent{\textbf{Title:}}\\ Intersection-based Ensemble for Small Multi-Object Tracking in Challenging Environments \\

\noindent{\textbf{Members:}}\\
\textbf{Guan-Zhang Wang$^1$ (guang.zhwa@gmail.com)}, Wei-Ta Chu$^1$ \\

\noindent{\textbf{Affiliations:}}\\
$^1$ National Cheng Kung University, Taiwan

\subsection{elsalabA}
\noindent{\textbf{Title:}}\\ Boosting Small Object Tracking via Collaborative Detection Transformer \\

\noindent{\textbf{Members:}}\\
\textbf{Bing-Cheng Chuang$^1$ \\(tars111060008@gapp.nthu.edu.tw)}, \\Jia-Hua Lee$^1$, Pin-Tseng Kuo$^1$, I-Hsuan Chu$^1$, Yi-Shein Hsiao$^1$, Cheng-Han Wu$^1$, Po-Yi Wu$^2$, Jui-Chien Tsou$^2$, Hsuan-Chi Liu$^2$, Chun-Yi Lee$^2$, Yuan-Fu Yang$^3$ \\

\noindent{\textbf{Affiliations:}}\\
$^1$ National Tsing Hua University, Taiwan\\
$^2$ National Taiwan University, Taiwan\\
$^3$ National Yang Ming Chiao Tung University, Taiwan

\subsection{sgm11}
\noindent{\textbf{Title:}}\\  Confidence-based Adaptive Weighted Boxes Fusion for Multi-Object Tracking of Small Birds \\

\noindent{\textbf{Members:}}\\
\textbf{Kosuke Shigematsu$^1$ (k-shigematsu@oita-ct.ac.jp)}, Asuka Shin$^1$ \\

\noindent{\textbf{Affiliations:}}\\
$^1$ National Institute of Technology, Oita College, Japan

\subsection{xmba15}
\noindent{\textbf{Title:}}\\ Multi-Object Bird Tracking via CenterNet Ensemble and Motion-Compensated OC-SORT \\

\noindent{\textbf{Members:}}\\
\textbf{Ba Tran$^1$ (thba1590@gmail.com)} \\

\noindent{\textbf{Affiliations:}}\\
$^1$ Axelspace Corporation, Japan

\section{Resources}
\label{sec:apd:resource}

The following resources are provided to support the SMOT4SB challenge and facilitate further research:

\begin{itemize}
  \item \textbf{Challenge website:} \url{https://mva-org.jp/mva2025/index.php?id=challenge}
  \item \textbf{Challenge platform (CodaBench):} \url{https://www.codabench.org/competitions/5101/}
  \item \textbf{Baseline code repository:} \url{https://github.com/IIM-TTIJ/MVA2025-SMOT4SB}
  \item \textbf{Dataset (Google Drive):} \url{https://drive.google.com/drive/u/1/folders/1Y1J13W6VlgDh-L28n_mVbs7HIfo_Hv5s}
\end{itemize}

\end{document}